\documentclass[conference]{IEEEtran}
\IEEEoverridecommandlockouts
\usepackage{cite}
\usepackage{amsmath,amssymb,amsfonts}
\usepackage{algorithmic}
\usepackage{booktabs}
\usepackage{graphicx}
\usepackage{multirow}
\usepackage{multicol}
\usepackage{textcomp}
\usepackage{xcolor}
\usepackage{subfigure}
\usepackage{float}
\usepackage{authblk}

\makeatletter
\newcommand{\linebreakand}{%
  \end{@IEEEauthorhalign}
  \hfill\mbox{}\par
  \mbox{}\hfill\begin{@IEEEauthorhalign}
}
\makeatother

\def\BibTeX{{\rm B\kern-.05em{\sc i\kern-.025em b}\kern-.08em
    T\kern-.1667em\lower.7ex\hbox{E}\kern-.125emX}}
\begin{document}

\title{SOLIS: Autonomous Solubility Screening using Deep Neural Networks\\
\thanks{This work was supported by the Leverhulme Trust through the Leverhulme Research Centre for Functional Materials Design and the H2020 ERC Synergy Grant Autonomous Discovery of Advanced Materials under grant agreement no. 856405. {Correspondence should be directed to \textit{gabriella.pizzuto@liverpool.ac.uk}}.}
}



\author[1]{Gabriella Pizzuto}
\author[2]{Jacopo de Berardinis}
\author[1]{Louis Longley}
\author[3]{Hatem Fakhruldeen}
\author[1, 3]{Andrew I. Cooper}
\affil[1]{Department of Chemistry, University of Liverpool, United Kingdom}
\affil[2]{Department of Informatics, King's College London, United Kingdom}
\affil[3]{Leverhulme Research Centre for Functional Materials Design, University of Liverpool, United Kingdom}

\maketitle

\begin{abstract}
Accelerating material discovery has tremendous societal and industrial impact, particularly for pharmaceuticals and clean energy production. 
Many experimental instruments have some degree of automation, facilitating continuous running and higher throughput.
However, it is common that sample preparation is still carried out manually.
This can result in researchers spending a significant amount of their time on repetitive tasks, which introduces errors and can prohibit production of statistically relevant data. 
Crystallisation experiments are common in many chemical fields, both for purification and in polymorph screening experiments. 
The initial step often involves a solubility screen of the molecule; that is, understanding whether molecular compounds have dissolved in a particular solvent. 
This usually can be time consuming and work intensive. 
Moreover, accurate knowledge of the precise solubility limit of the molecule is often not required, and simply measuring a threshold of solubility in each solvent would be sufficient.
To address this, we propose a novel cascaded deep model that is inspired by how a human chemist would visually assess a sample to determine whether the solid has completely dissolved in the solution.
In this paper, we design, develop, and evaluate the first fully autonomous solubility screening framework, which leverages state-of-the-art methods for image segmentation and convolutional neural networks for image classification.
To realise that, we first create a dataset comprising different molecules and solvents, which is collected in a real-world chemistry laboratory.
We then evaluated our method on the data recorded through an eye-in-hand camera mounted on a seven degree-of-freedom robotic manipulator, and show that our model can achieve 99.13\% test accuracy across various setups, while being simple and fast to train and, as a result, easily transferable to a robotic platform.

\end{abstract}


\section{Introduction}
Laboratory robotics working collaboratively with researchers using modular and reconfigurable lab equipment brings us closer to the vision of Industry 5.0~\cite{Demir2019}, which has the overarching goal of adopting human-centric approaches to digital technologies. 
Self-driving laboratories that are capable of planning, executing and evaluating the results of experiments have the potential to make faster, more efficient progress in materials research.
The automation specific devices, such as sample changers, is generally confined to a specific task and has a high cost of entry, making it inaccessible to many research groups. 
Also, autonomous labs can only be realised through linking the abstract concepts of chemical processes with the hardware responsible for its execution~\cite{Bai2021}, making use of algorithms to facilitate decision making for scientists. 
These algorithms should be robust to real laboratory conditions and importantly, generalisable to different experiments. 
This would allow scientists to trust the predictions and in turn facilitate automation of chemical discovery.

Currently, human scientists spend a significant amount of time on mundane tasks because robotic platforms are only capable of performing a limited set of operations in comparison to humans, and most of these require the scientist to do all of the decision making, even for the most trivial scenarios~\cite{Shiri2021}. 
Hence, a long-term goal is to reduce both human labour with respect to tedious tasks and to have algorithms that carry out mundane resolutions, bringing us closer to a digitalisation of chemical manufacturing which is a critical technology path toward a sustainable society~\cite{Inderwildi2020}. 

\begin{figure}[t!]
    \centerline{\includegraphics[width=0.35\textwidth]{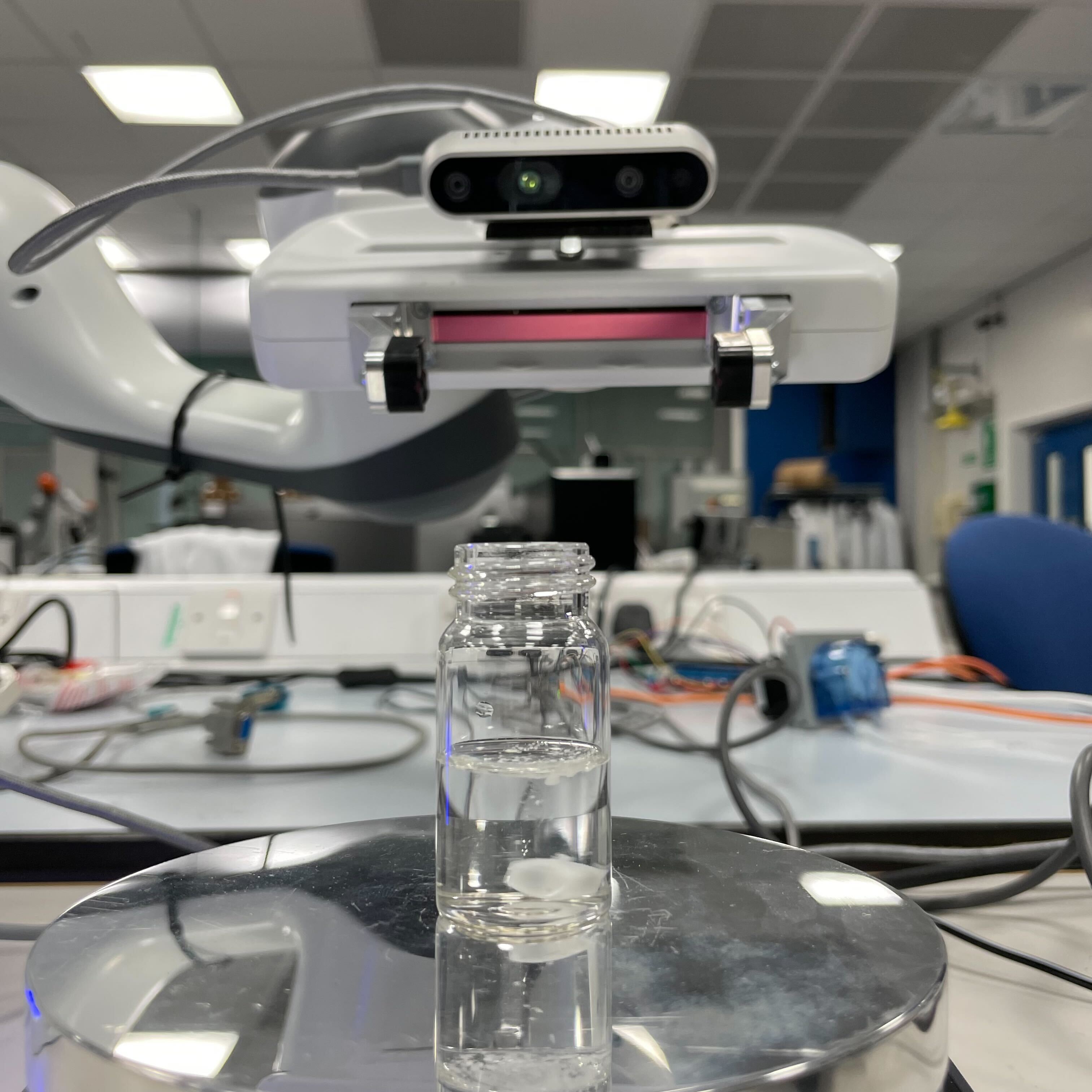}}
    \caption{The robotic setup used for our model architecture to be trained on and evaluated for solubility screening.}
    \label{fig:setup}
\end{figure}

Material discovery laboratories rely on diverse experimental workflows that are key to development of new, sustainable materials.
An important process in such laboratories is solubility screening. 
Solubility screens can be performed in a variety of different ways, such as by titration, ultraviolet-spectroscopy or high performance liquid chromatography (HPLC) methods~\cite{Petereit2011}, all of which involve expensive, customised equipment. 


A recent method has evaluated low-cost equipment (cameras) with a simple computer vision method that measures turbidity~\cite{Shiri2021}, where the captured image is analysed to measure the average brightness of the solute-solvent mixture. 
The latter is then compared to a reference image of the solvent to conclude whether the solution is fully dissolved or not. 
The brighter the image, the more undissolved solute is present because the clear solvent appears black when placed against a dark background. 
However, this approach is limited to simple conditions within highly controlled environments that may not be easily generalisable in complex real-world laboratory environments. 
Specifically, initial researcher input is required every time the workflow runs, a black background is needed, and the method does not tolerate different lighting conditions. 

Recently, vision methods based on deep neural networks and convolutional neural networks (CNNs) have proven effective in laboratory and medical environments~\cite{Eppel2020}.
Such methods have gained traction for visual recognition of lab hardware~\cite{Eppel2022} and have shown promising results in challenging real-world settings.
This is remarkable since laboratory glassware and many solvents are in fact transparent and hence pose a significant challenge for classical machine vision algorithms.
Deep models pave the way for future laboratory robotics in autonomous laboratories and will be key to overcome current limitations in methods relying on more traditional approaches.

Inspired by how human chemists visually assess solutions, we propose a novel optimal solubility screening method that does not require analytic instruments, such as HPLC-based methods, and that by using deep models as feature extractors can generalise to standard laboratory environments. 
Since accurate knowledge of the precise solubility limit of a molecule is often not required, it can be enough to measure a threshold of solubility in a solvent, for example to facilitate solvent-antisolvent polymorph screening.
As illustrated in Fig.~\ref{fig:setup}, our novel model architecture for solubility screening is evaluated in a real-world laboratory with all the challenges associated with this environment. 
The input images are recorded directly from the Franka Emika Panda robot and these are used to train and evaluate our cascaded system combining the state-of-the-art mask regional-CNN (R-CNN), which feeds the features from the cropped region of interest (RoI) around the sample vial into the classifier trained to determine whether the solute-solvent has fully dissolved.

In summary, our main contributions in this paper include:
\begin{enumerate}
    \item the first cascaded deep model for solubility screening, relying on state-of-the-art image segmentation and classifier;
    \item an end-to-end architecture without human intervention or pre-calibration;
    \item the evaluation of the proposed model on a novel dataset collected in a a conventional, unmodified materials chemistry laboratory.
\end{enumerate}

The rest of the paper is structured as follows: Section~\ref{sec:related_work} gives a review of related work on solubility screening, and Section~\ref{sec:methodology} reports the novel computational model for automated solubility screening and the dataset collected. 
Section~\ref{sec:exp_eval} details the experimental evaluation, including
the results obtained. 
Finally, Section~\ref{sec:conclusion} draws conclusions and gives direction for future work.

\section{Related Work}
\label{sec:related_work}
Solubility screening has been used in research and development laboratories for a wide array of experimental studies and it remains a key process in many workflows.
In most cases, the process is a case of adding `excess solid' or `excess solvent'. 
For the former, different analytical methods can be used once the solution reaches equilibrium, such as titration techniques, gravimetric methods, HPLC, or NMR~\cite{Black2013}. 
Another method for the `excess solid' process involves filtering, drying, and weighing the undissolved solid~\cite{Black2013}.
On the other hand, for the `excess solvent' process the common practice is to add solvent to the solute until the latter fully dissolved in it~\cite{Black2013}. 
In this case, techniques such as Fourier transform infrared spectroscopy (FTIR)~\cite{daSilve2011} are used to measure the point at which the solute dissolves.
While both methods have their respective strengths and shortcomings, the `excess solvent' process is best suited for integrating with robotic platforms as it requires less specialist equipment, and does not require additional filtration/separation steps. 
It also happens to be the one that in practice is most widely used. 

Given that the method of using a light source and a detector aims at replicating the human eye, methods relying on measuring ‘turbidity’ or a ‘clear point’ were introduced.
A recent work proposed a method to measure the brightness of the solution of captured images~\cite{Shiri2021}. 
While this method uses low-cost and non-invasive equipment, the main challenge that it fails to address is robustness to dynamic laboratory environments and being an end-to-end model that does not require human input.
Moreover, as the method relies on the user selecting the RoI, an inexperienced scientist might not select the correct area and in turn this would affect the results.
In fact, while here the RoI is is a manually determined slice of the vial that would vary depending on the human chemist/experiment, our method uses the whole vial and hence is the same every time.

Our proposed approach addresses this by adopting the mask R-CNN to focus on the RoI, and hence remove the need of human input. 
Together with using deep models for extracting features from the RoI, our method provides flexibility in using the network on images without the need of adding a black background or fixing the illumination.
From the technical point of view, our approach is novel because the proposed framework combines mask R-CNN with deep model classifiers for decision-making within a laboratory environment, which is still an open problem.
Our architecture is able to deal with data coming from images recorded either from the human chemist or the ‘robotic chemist' due to its ability in dealing with background and illumination variance.

\begin{figure*}[ht]
    \centerline{\includegraphics[width=0.99\textwidth]{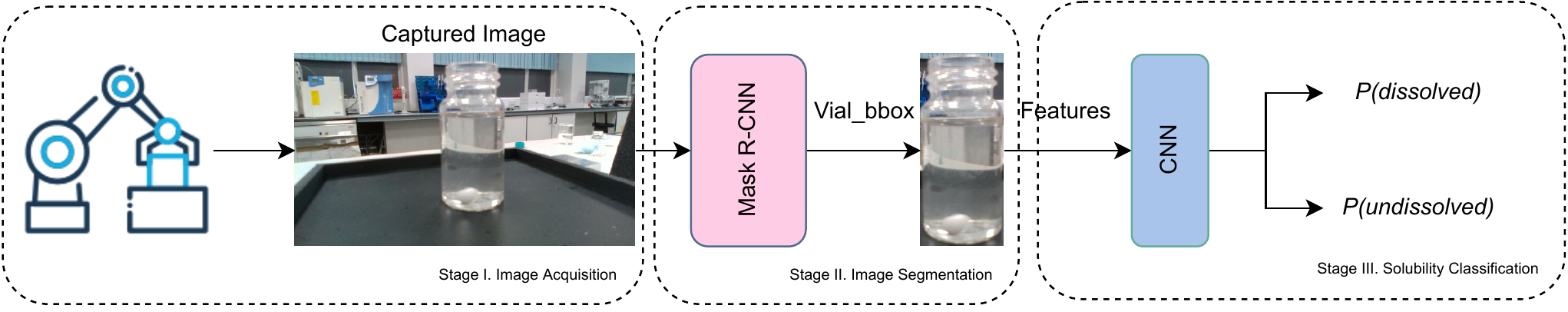}}
    \caption{Overview of the proposed approach for determining whether a molecule dissolves in a given solvent. An input image recorded from the Franka Emika Panda robot is fed into our cascaded system combining the mask R-CNN, which feeds the features from the cropped RoI around the vial into the classifier trained to determine whether the solute-solvent has fully dissolved.}
    \label{fig:overall_block_diagram}
\end{figure*}

\section{SOLIS Architecture} 
\label{sec:methodology}

Our model here aims at addressing the challenge of determining which molecules dissolve in a given solvent. 
Our primary goal and main interest is to obtain an architecture that can be transferred to a robotic platform in a laboratory environment that has been designed by and for human chemists. 
We present our cascaded system combining a mask R-CNN, with a deep model classifier for solubility screening. 
An overview of the proposed method is illustrated in Fig.~\ref{fig:overall_block_diagram}. 
Section~\ref{sec:cascaded_model} gives an in-depth overview of the cascaded model, while further details about the dataset we collected to evaluate our models is given in Section~\ref{sec:dataset}.

\subsection{Cascaded Model}
\label{sec:cascaded_model}

Our system relies on the state-of-the-art CNNs tailored for autonomous solubility screening. 
The architecture can detect glass laboratory vials captured directly from eye-in-hand camera mounted on the robotic manipulator, without the need of a human scientist to highlight the RoI.
Given the coordinates of the bounding box predicted around the mask of the vial, we use the features from this region to determine whether the molecule has dissolved.
As shown in Fig.~\ref{fig:overall_block_diagram}, our architecture is mainly divided into three stages.
In the first part, we collect and preprocess the image.
Following this, the data is fed into an image segmentation model and a CNN-based image classifier.
Our system can be deployed on a robotic platform as a ``robotic scientist'' to help lab-based human scientists to avoid repetitive tasks.

\subsubsection{Image Acquisition}
\label{subsubsec:image_acquisition}
For the image acquisition stage, we use a Franka Panda Emika robot equipped with an Intel RealSense D435i camera on its end-effector; the latter attached via a 3D printed mount. 
The robot perceived its workspace from above the IKA Plate, as illustrated in Fig.~\ref{fig:setup}.
All images were recorded using a framerate of 30 frames per second.
In contrast to~\cite{Shiri2021}, we opted to put the camera on the end-effector of the robot rather than as its own station since this gives us the possibility to use the camera also for scene understanding and visual servoing.  
It also offers the future potential to mount this camera on a mobile ``robotic chemist''~\cite{Burger2020}. 
For these instances, the mask R-CNN from the first stage could easily be adapted to realise such tasks.

\subsubsection{Image Segmentation} 
\label{subsubsec:maskrcnn}
Following successful image acquisition, the next step is to extract meaningful features for training the solubility classifier. 
As a result, the first stage of the cascaded model is a mask R-CNN~\cite{He2017}. 
Given that the mask R-CNN is currently the state-of-the-art method for instance segmentation with respect to a large array of image segmentation tasks~\cite{He2017}, this work adopts the pre-trained model on the TransProteus dataset~\cite{Eppel2022}.
The role of this model is to find the region and boundaries of the glass laboratory vial in the image that is recorded directly from the end-effector of the Panda Emika Franka robotic arm. 
As we are operating in a challenging environmental setup, it would be ideal for the classifier to focus on the output segment that is contained within the RoI mask. 
By taking this approach, there is still the important feature of limiting the region of the image in which the output segment is found while removing the need for human input in selecting the RoI as in related works~\cite{Shiri2021}.
Specifically for our task, the mask R-CNN uses ResNet as a backbone~\cite{He2016}. 
From the candidates proposed by the mask R-CNN, the instance bounding box and class are predicted by the box head. 
Here, we use $vial\_bbox$, which are the coordinates of the bounding box of the glass vial that are predicted by the mask R-CNN.
For the next step, the underlying image features obtained from the bounding box are used as input to the CNN.
While masks are also provided by the network for both the vessel (glass vial) and the material (solution), our work only requires the image segmentation to produce a RoI with useful features of the complete vial for the classifier; hence, we only used the bounding box around the segmented mask.

\subsubsection{Solubility Classification}
\label{subsubsec:solubility_classification}
Here, we view the problem of solubility screening as a classification process, since for many material discovery experiments, we only need a threshold of solubility. 
Convolutional neural networks are the current state-of-the-art methods for image classification.
Therefore, given the challenging data that is involved in our application, this framework was the obvious choice.
Given the small dataset, we opted to use transfer learning from a pre-trained network on the Imagenet dataset.
For a thorough evaluation, we looked into both finetuning and feature extraction strategies for training.
For finetuning, the pretrained model is retrained and hence all of its parameters, that is, the higher order features and the classifier are updated with respect to the current task of classifying whether the dissolved state has been reached.
For feature extraction, the model is frozen such that only the output layer was updated; hence, the feature maps from the pre-trained model are retained.
Another important criteria we considered was image augmentation, where we adopted random resizing, cropping, and horizontal flipping for the training set and resizing and center cropping for the validation set.
All the data was normalised prior to feeding it into the model architecture.
We have tested our cascaded model using different CNN architectures, and we give further detail of these setups in the experimental evaluation (Section~\ref{sec:exp_eval}).


\subsection{Dataset} 
\label{sec:dataset}

\begin{figure*}
    \centering
    \subfigure[]{\includegraphics[width=0.245\textwidth]{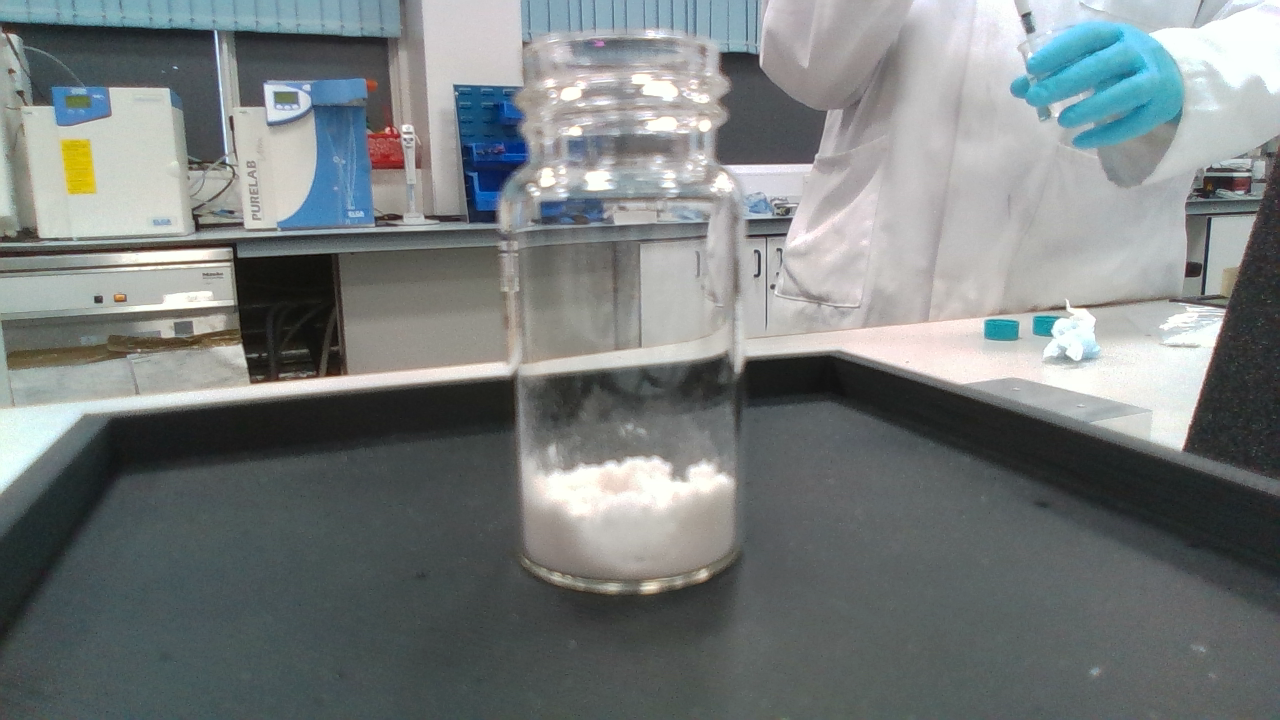}} 
    \subfigure[]{\includegraphics[width=0.245\textwidth]{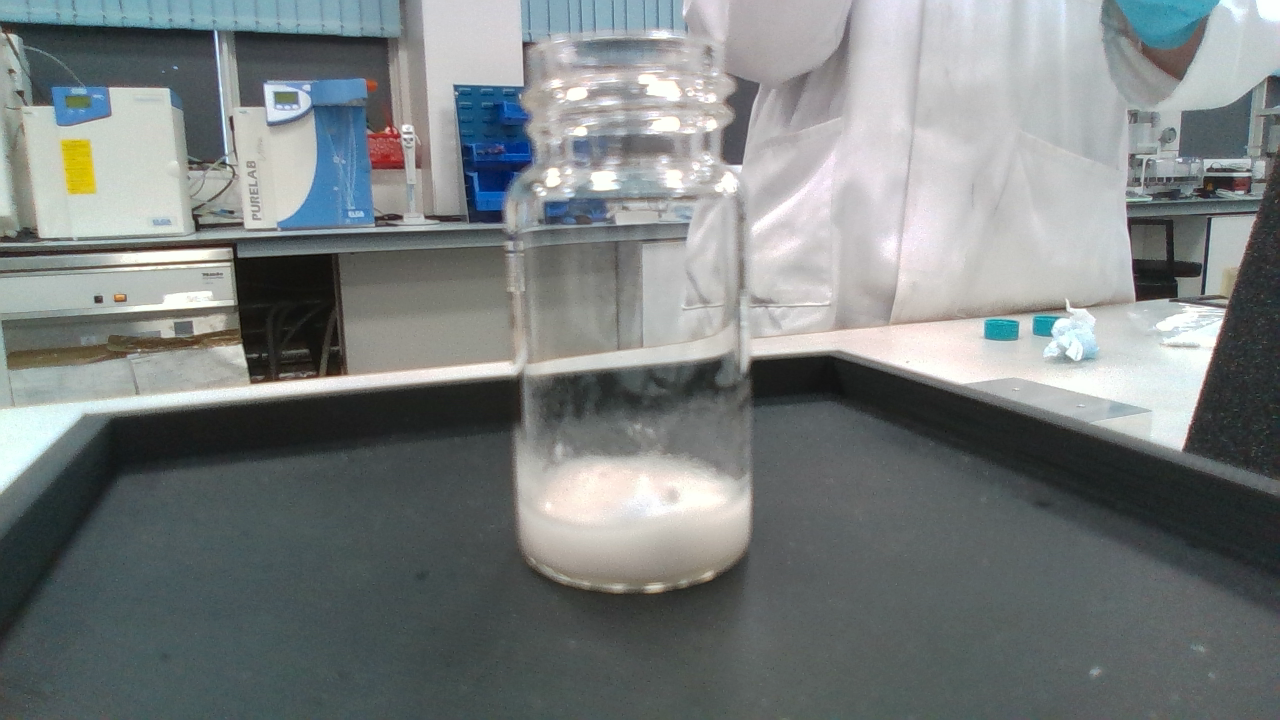}} 
    \subfigure[]{\includegraphics[width=0.245\textwidth]{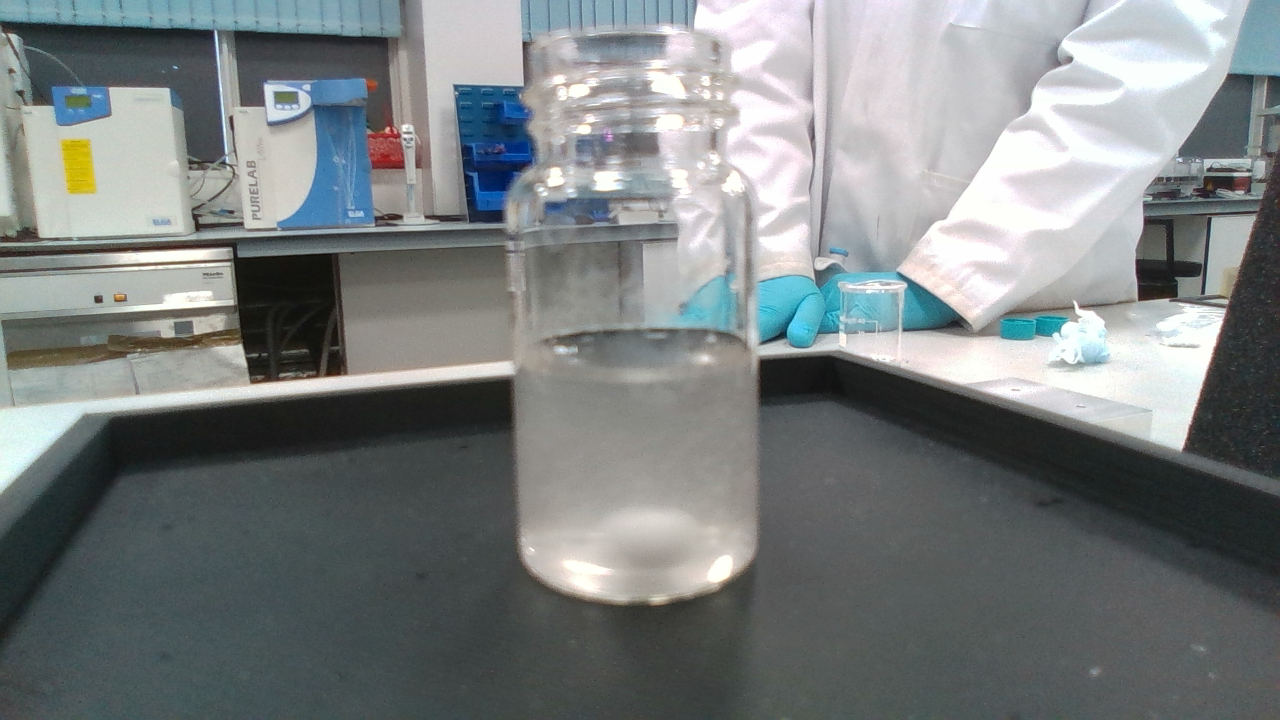}}
    \subfigure[]{\includegraphics[width=0.245\textwidth]{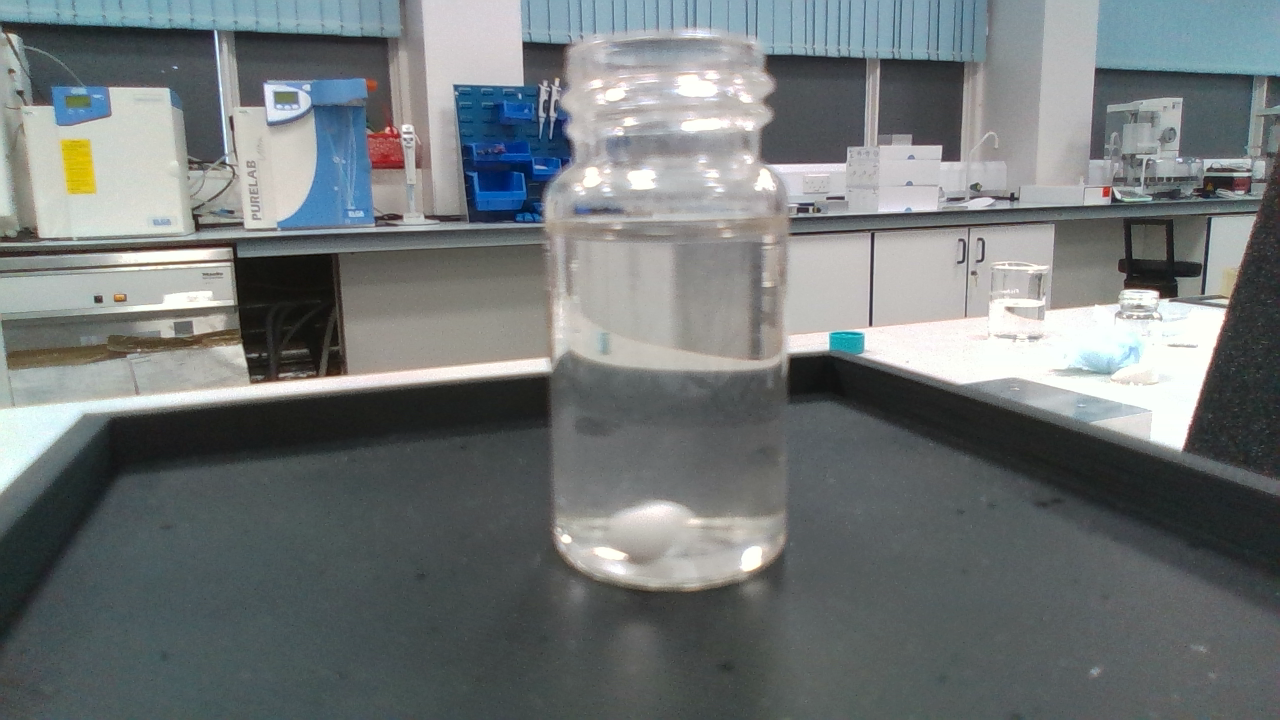}}
    \caption{An overview of the recorded dataset for benzimidazole with acetone. (a) Initial solid dispensing is carried out, where benzimidazole is weighed into the glass vial. A magnetic stir bar is also placed in the vial. (b) Initial solvent dispensing occurs, and the stirring plate is also switched on to mix the solution. (c) Mixing of the solution continues and the mixture becomes partially dissolved. (d) The solution becomes fully dissolved.}
    \label{fig:dataset}
\end{figure*}

Given the novelty of this application, there exists no dataset that is publicly available for automated solubility screening. 
Moreover, the images recorded in a relevant study~\cite{Shiri2021} were recorded using a black background and a reference vial.
While we too considered a setup with a black background, we specifically focused on having non-sterile conditions to facilitate easy deployment in a real laboratory environment.
Hence, it was crucial for us to build a specific dataset for this problem.
We collected data directly from a materials discovery laboratory where a human chemist manually carried out the ‘excess solvent' procedure.
After preparing the solute in 20 mL vials with a magnetic stir bar inside, the chemist pipetted the solvent into the sample using increments of initially 1 mL but then 0.5 mL when it was judged by eye that sample turbidity was dropping.
During the experiment, the sample was continuously stirred with a magnetic stirrer\footnote{IKA Plate (RCT digital) was used for our experiments.} to prevent the powder dropping to the bottom of the vial. 
Stirring also prevented sample clumping
The full dataset comprises 2 solids: caffeine and benzimidazole and 3 solvents: water, ethanol and acetone; hence six different setups were used.
The choice of these chemicals pivoted around the fact that these can be used safely on an open-bench and could be integrated in a crystallisation workflow. 
The experiments were supervised by a human chemist, who was also responsible for annotating the dataset.
Hence, the dataset collection process was terminated once the human chemist visually decided that the solute had dissolved.
As mentioned, the RGB images were recorded using an eye-in-hand camera on a Panda Emika robotic manipulator.

\section{Experimental Evaluation}
\label{sec:exp_eval}
We posed two questions to evaluate this method.
First, how does SOLIS (our proposed approach) perform on the autonomous solubility screening task?
More precisely, this first experiment aims to compare the performance of different state-of-the-art artificial neural networks to implement the CNN classifier\footnote{We generally refer to this part of the pipeline as the CNN classifier, regardless of the specific architecture implementing it. While the previous stage, the mask R-CNN, is also a CNN classifier, we usually refer to this as the image segmentation part.} in our pipeline (\textit{c.f.} Fig.~\ref{fig:overall_block_diagram}).
The criteria for this comparison are the average classification performance on the solubility screening task and the complexity of the classifier (number of parameters).
These are both significant for the selection of the specific classification model that will be deployed in SOLIS, as we aim at minimising inference times and the risk of failure.
Second, given that solubility screening is usually integrated within a larger workflow, when does it fail?
From the first experiment, we identify the most accurate and parsimonious model to deploy in SOLIS, which is then used in the second experiment.
This comparison aims to illustrate that, although SOLIS can produce misclassifications, our understanding of when this happens will equip us with the right approach to compensate for these predictions and, as a result avoid, interrupting expensive and time-consuming experiments.

\subsection{Experimental Setup}
For both the experiments, we evaluated all models on the solubility screening task, quantifying their ability to predict the state of the solution given a raw image recorded directly from the robotic platform.
This was done by computing the cross-entropy loss and accuracy of each model on the test set.
Cross entropy (CE) is the dominant measure for training and evaluating classification models.
Intuitively, it measures the difference between two probability distributions: the target distribution (the actual solubility status), and the predicted one (the model's predictions of the solubility status).
Accuracy is derived from the model's predictions by considering the class associated with the maximum activation for a given input.

The framework is powered by PyTorch, using a machine equipped with an AMD Ryzen Threadripper 3970X 32 Core CPU and a single NVIDIA GeForce RTX 3090 Graphical Processing Unit (GPU).

\begin{table}
\centering
\caption{Summary of the CNN architectures tested as classifiers for solubility screening. The number of trainable parameters makes distinction between the backbone of each architecture and the last fully-connected layer used for classification (in brackets).}
\label{tab:cnn-summary}
\begin{tabular}{@{}lll@{}}
\toprule
CNN model & Input size & No. of parameters \\ \midrule
VGG         & 224x224    & 128 M (8.2K)      \\
ResNet18    & 224x224    & 11.2 M (1K)       \\
InceptionV3 & 224x224    & 24.3 M (5.6K)     \\
Densenet    & 299x299    & 7 M (2K)          \\ \bottomrule
\end{tabular}
\end{table}

\subsection{Solubility Screening Classification}

In this section, we explore different networks for the solubility classifier.
The architectures that were used include VGG~\cite{Simonyan15}, ResNet~\cite{He2016}, Inception~\cite{Szegedy2016} and Densenet~\cite{Huang2017}.
The input size expected by each architecture and the overall complexity are reported in Table~\ref{tab:cnn-summary}.
Although there are different models that could have been applicable,  our decision was also motivated by having an architecture that has proved to work in different domains and hence using the reference fields to address our challenging task was the most sensible approach.
Moreover, given that the method presented in~\cite{Shiri2021} requires human input, we could not directly compare this approach to ours since our proposed does not require human interventions.

As the RoI from the output of the mask R-CNN varied, we scaled the features to the input size each architecture expects (\textit{c.f.} Table~\ref{tab:cnn-summary}).
Nonetheless, a different transformation strategy was followed for the training sets in order to add stochasticity during training.
In particular, for each epoch, we select a random region from the resulting segmentation -- matching the input size of the classifier.
Although this approach may result in some samples where the solubility status is hard to estimate/detect (\textit{e.g.,} the liquid appears only in a relatively small area), we expect the regularising effect to improve the generalisation capabilities of the resulting models.

All CNN models had been pre-trained on Imagenet~\cite{Deng2009}, then considered for transfer learning on the solubility screening task.
To achieve this, the last fully-connected layer of each architecture was discarded (as it was originally designed for ImageNet's classes) and replaced with a linear fully-connected layer with 2 output units - one per solubility outcome (\textit{undissolved}, and \textit{dissolved}).
Each CNN classifier was then trained to minimise the CE loss according to two different strategies: \textit{feature extraction}, where the gradient is back-propagated only up to the last (new) fully-connected layer - meaning that the parameters of the rest of the network (the backbone) are left unchanged, and \textit{fine-tuning}, where the gradient is fully back-propagated and all the parameters of the network are adjusted accordingly.

We used the Adam optimiser with a learning rate equal to 0.001, together with an early stopping policy (with 10 epochs of patience, and a delta of $10^{-2}$) to prevent over-fitting.
Considering the limited size of the dataset, we performed 5-fold cross validation.
This allowed to evaluate each architecture without exposing our results to the potential bias of a single (fixed) split.
The dataset was split into 5 partitions (folds) and each of them was used as the test set of a distinct experiment (for the same architecture).
The predictions of models on each test fold are then aggregated, so that test losses and accuracy can be computed.
Although the splits were produced apriori for all the experiments, we also set a random seed (1992) for a number of Python modules (\texttt{numpy}, \texttt{pandas}, \texttt{torch}, \texttt{math}) to ensure reproducibility.

The results are presented in Table~\ref{tab:average accuracy across all models}.
All of the models give an overall good performance that would be suitable to deploy in laboratory settings.
However, the best performing models are the ones that are finetuned.
When looking at the test loss, ResNet18 achieves the best performance in terms of average CE loss and standard deviation. 
We believe that although transfer learning aided our training methods as the dataset was relatively small (as the task was quite unique and the images are very different to the ones found in Imagenet) training only the last fully connected layer was not enough in all models.
Hence, the finetuning approach, while requiring higher training times in comparison, achieved the best performance.

\begin{table}
\caption{Overview of the performance of our proposed architecture with the CNN classifiers. For each model, we use finetuning for training the network and this is represented as FT, and we also used feature extraction during training which is illustrated through FE.}
\begin{center}
\begin{tabular}{@{}llll@{}}
\toprule
Model           & Method & Test Accuracy $\uparrow$ & Cross Entropy Loss $\downarrow$ \\ \midrule
VGG       & FT & 0.9846        & 0.0694 \textpm~0.6645    \\
VGG        & FE & 0.9675        & 0.0872 \textpm~0.3853    \\
ResNet18    & FT & 0.9904        & \textbf{0.0264 \textpm~0.2486}    \\
ResNet18    & FE & 0.9483        & 0.1274 \textpm~0.4345    \\
InceptionV3 & FT & \textbf{0.9913}        & 0.0295 \textpm~0.3989    \\
InceptionV3  & FE & 0.9334        & 0.1759 \textpm~0.4137    \\
Densenet  & FT& 0.9911        & 0.0363 \textpm~0.5648    \\
Densenet   & FE & 0.9528        & 0.1188 \textpm~0.4177    \\ \bottomrule
\end{tabular}
\label{tab:average accuracy across all models}
\end{center}
\end{table}

\subsection{Robustness Evaluation}
For our application, solubility screening requires a high level of accuracy to have a successful completion of the laboratory experiment. 
For example, solubility screening is a precursor for widely used pharmaceutical applications, such as crystallisation. 
Hence, should the solubility screen produce incorrect results this could be costly.
As a result, understanding the cases where it could fail is important to ensure workflow success and increase robustness of the overall system.

\begin{figure}
    \subfigure[]{\includegraphics[trim={0 2.5cm 0 5cm},clip, width=\columnwidth]{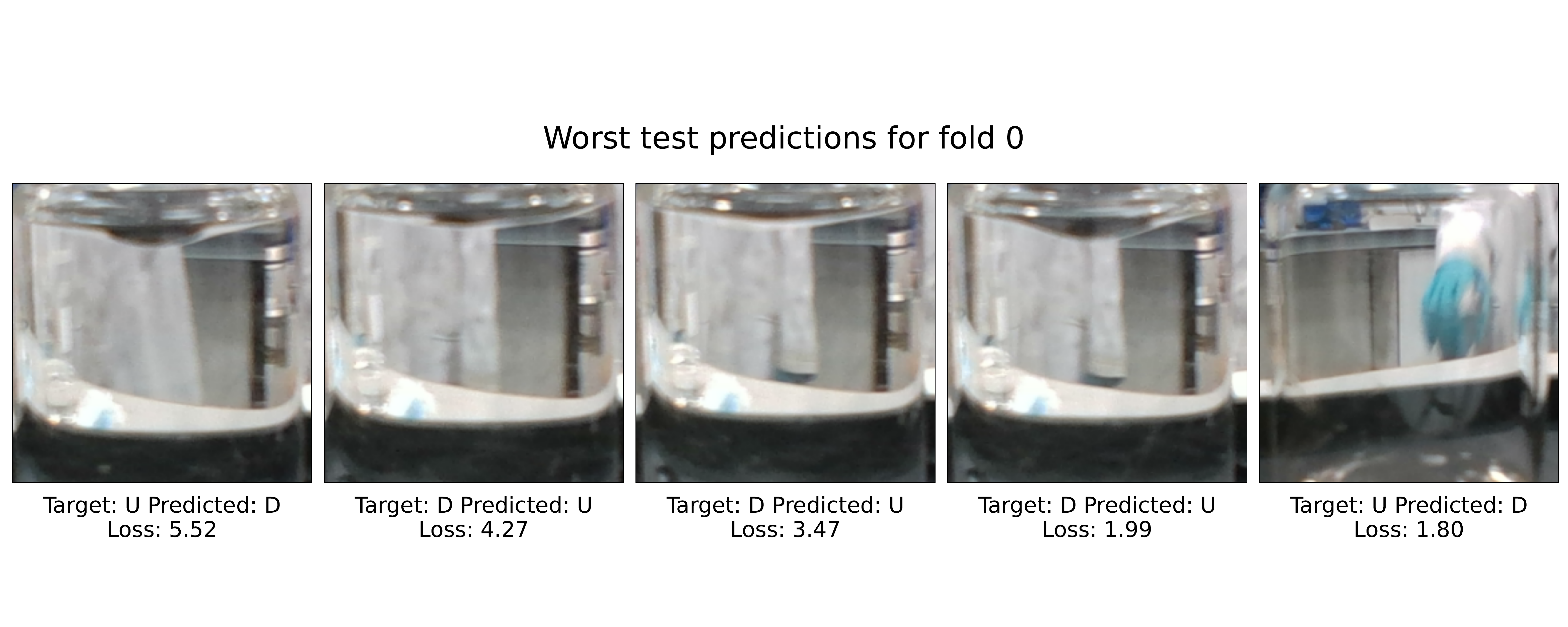}} 
    \subfigure[]{\includegraphics[trim={0 2.5cm 0 5cm},clip,width=\columnwidth]{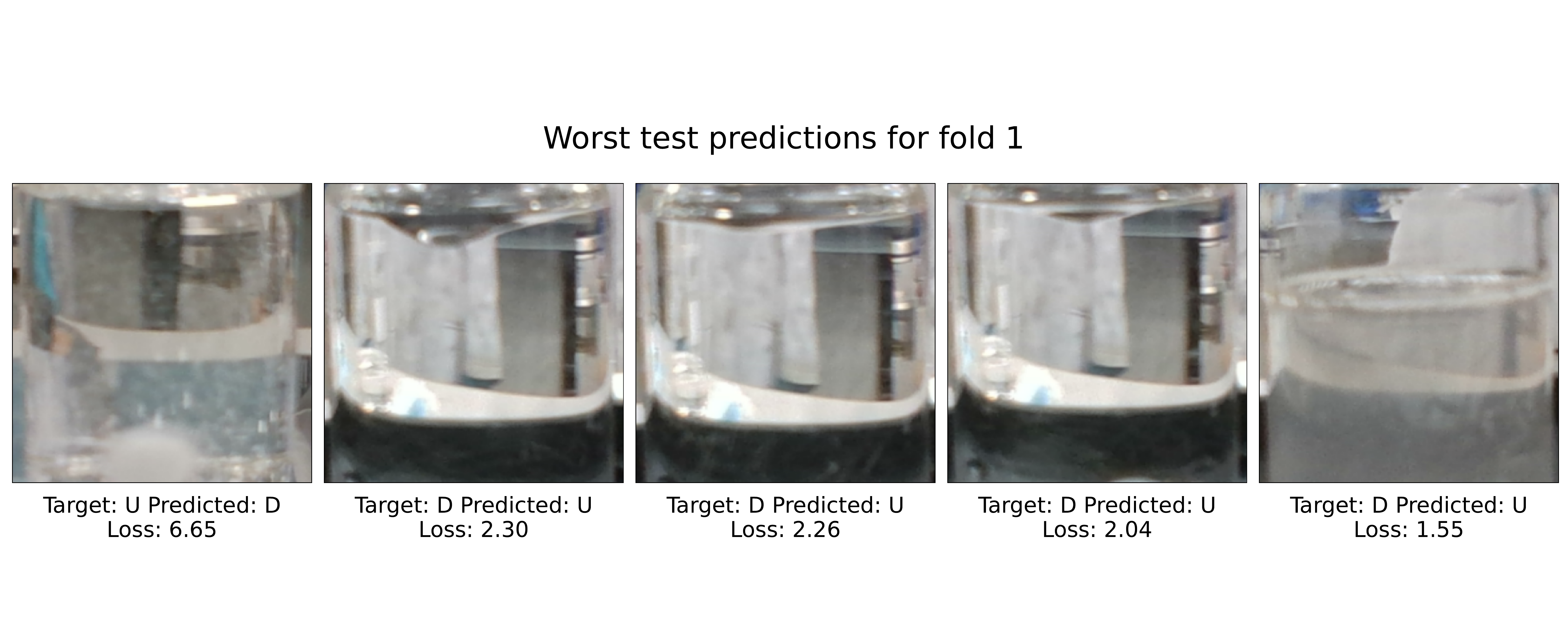}}  
    \subfigure[]{\includegraphics[trim={0 2.5cm 0 5cm},clip,width=\columnwidth]{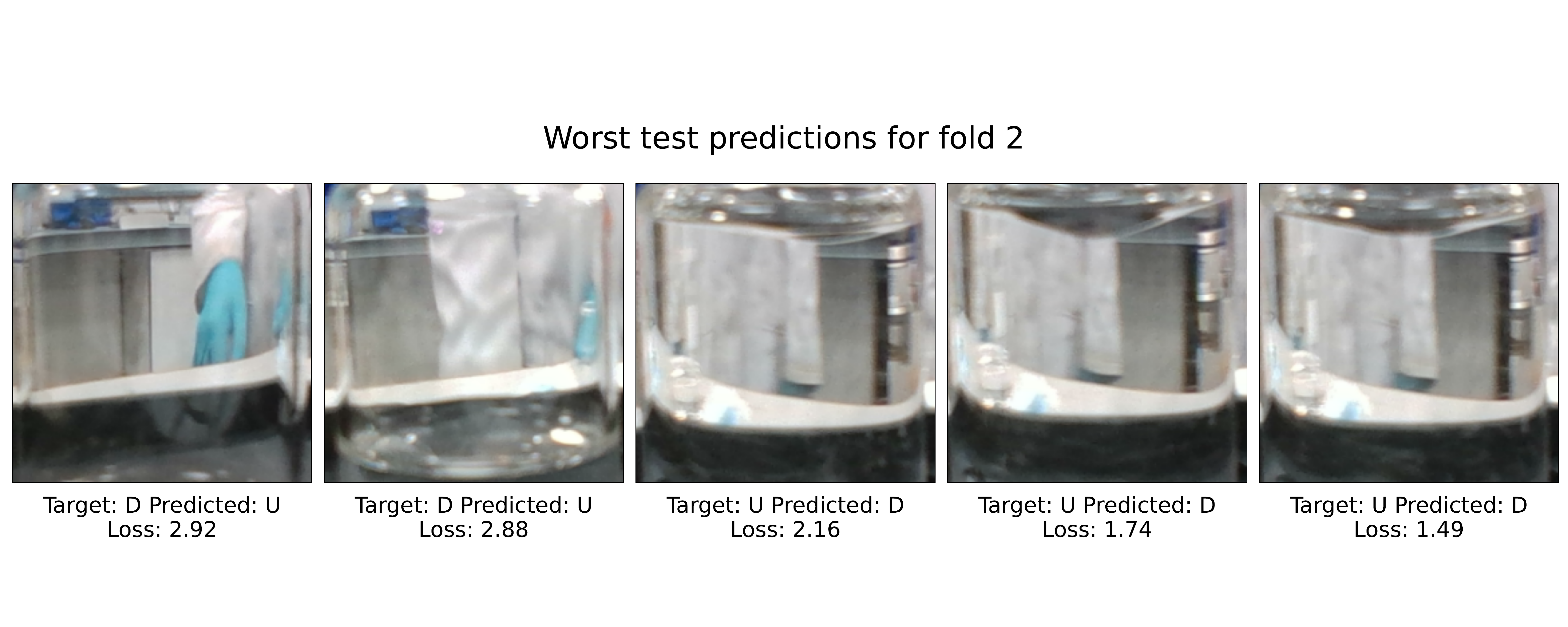}} 
    \subfigure[]{\includegraphics[trim={0 2.5cm 0 5cm},clip,width=\columnwidth]{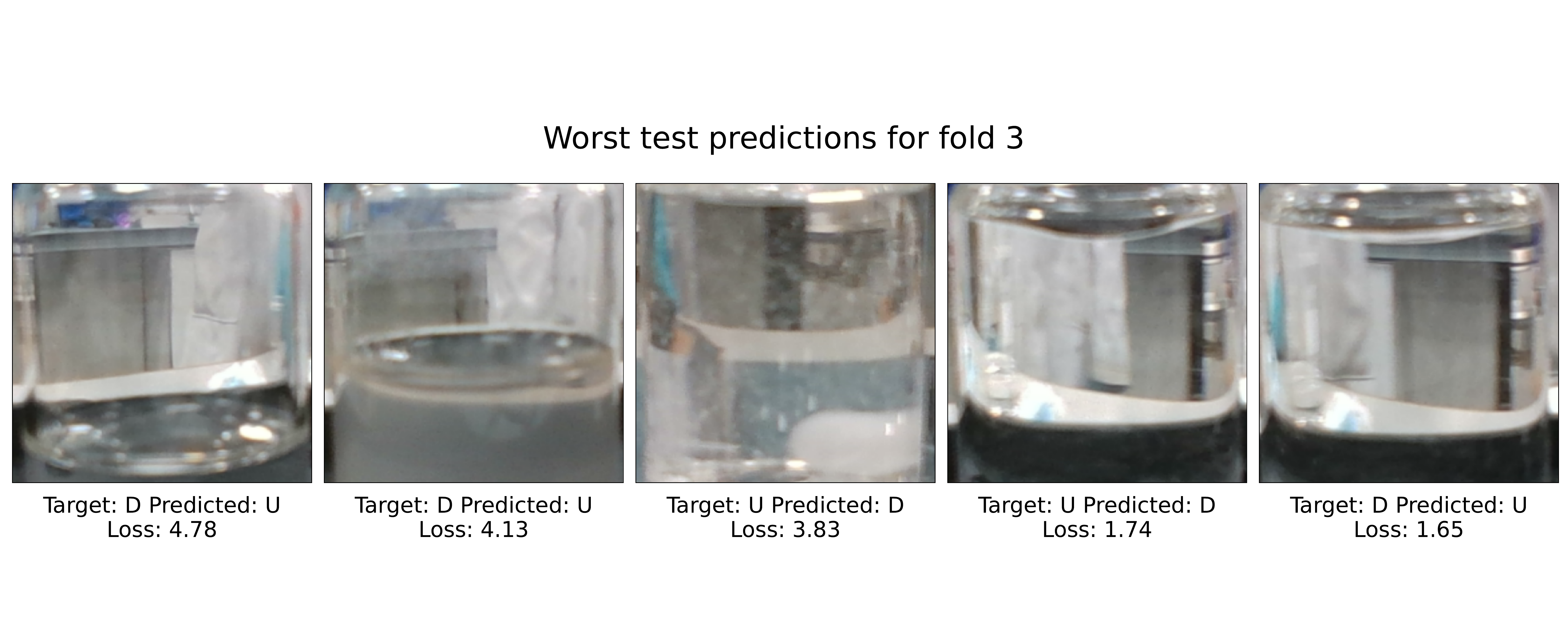}}
    \subfigure[]{\includegraphics[trim={0 2.5cm 0 5cm},clip,width=\columnwidth]{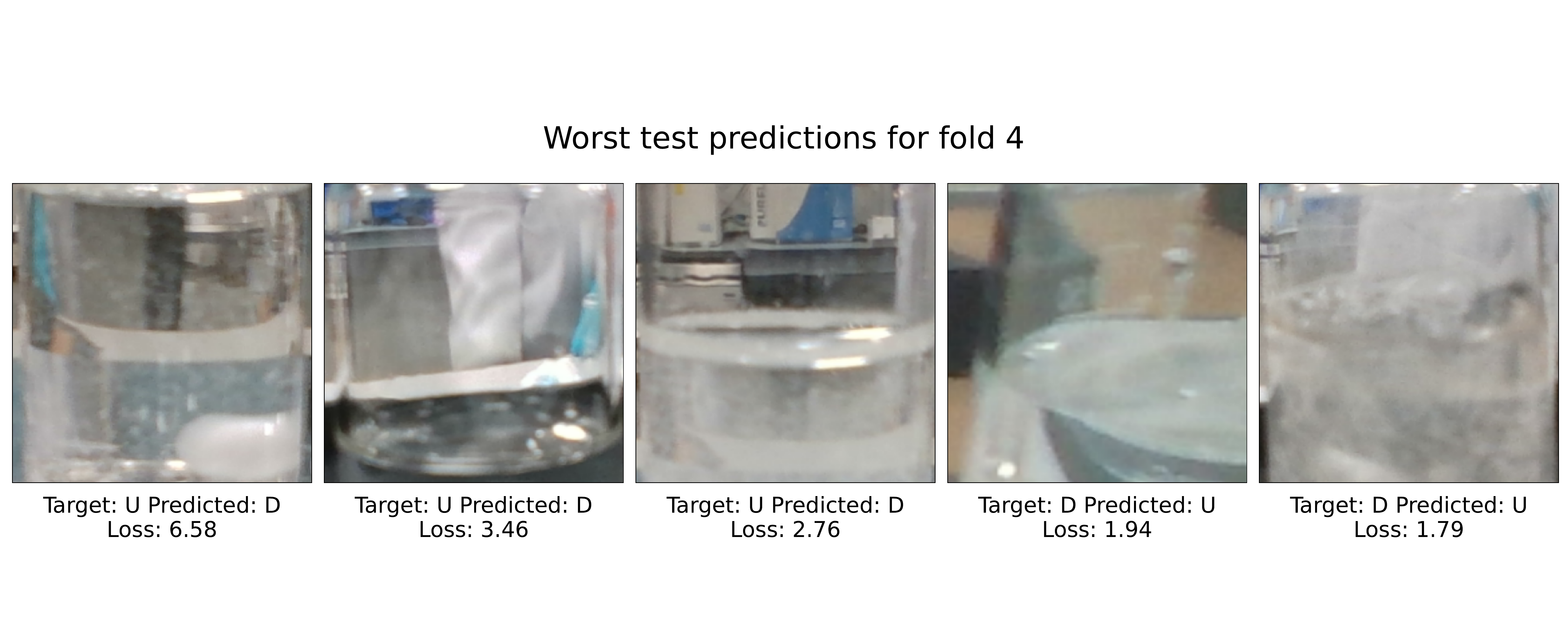}}  
    \caption{An in-depth analysis of the misclassifications across the five different folds for the ResNet18 model, which is finetuned. For each fold, we report the top five worst predictions.}
    \label{fig:prediction_errors}
\end{figure}

Since we opted for ResNet18 as the final model for deployment, we analysed in-depth the misclassifications and these are illustrated in Fig.~\ref{fig:prediction_errors}. 
Fig.~\ref{fig:prediction_errors} illustrates the five worst performances for each cross-validation fold.
At a closer look, we believe that a large proportion of the errors, for example Fig.~\ref{fig:prediction_errors}.b (1) and (V), result from reflections or spots on the glass vial walls.
In such cases, even a human chemist might misclassify the solution and to mitigate these errors they would take another look at the sample, potentially in different settings or under different light conditions.
Motivated by this, our future plan is that when the model predicts the dissolved state, we create a buffer of images rather than use a single one to base our decision on.

Another issue is that at the early stages of the experiment, the vial does not have a large proportion of solution and if the majority of the crop takes into account the top part of the transparent glass vial, it would be easy for the network to confuse this with a clear dissolved liquid.
Future work could also explore the use of a third classification category indicating cases that are borderline between dissolved and undissolved. 
This could be used as a basis for subsequent experiments with the same solvent and solute molecule aiming at more accurate determination of solubility limits.

To mitigate the challenge this brings (as illustrated in some cases in Fig.~\ref{fig:prediction_errors}), in future work we plan to explore how different image crops can affect the overall performance of our model.

\subsection{Discussion}
\label{subsec:discussion}
All of our experiments were carried out using the recorded dataset (Section~\ref{sec:dataset}). 
While we emphasise that the dataset was carried out in real-world laboratory conditions, in our preliminary experiments we explored the option of including the same black background, as in~\cite{Shiri2021}, and also went further to explore different lighting conditions.
In our case, as the mask R-CNN in the first step puts attention on the glass vial through the RoI selected within the bounding box, this step made our model robust to different environmental conditions and, as a result, we did not observe a better performance when using a black background.
The use of the mask R-CNN in the first stage also gives the possibility of having the vial in different positions, not necessarily within the centre of the image, and the proposed pipeline would still work.  
Again, this is quite advantageous as it never is the case that human or robotic scientists place glassware exactly in the same location.

Our long-standing goal is to have an end-to-end model that is not only accurate, but is also computationally efficient for deployment on a real robot.
As such, we remained faithful towards models that have a smaller number of parameters (such as opting for a ResNet18 against a ResNet152).
From the results obtained in Table~\ref{tab:average accuracy across all models}, it is evident that the difference between the finetuned models ResNet18, InceptionV3 and Densenet is very small.
In fact, it is our understanding that when deployed on a robotic platform all of these models would perform similarly.
Therefore, knowing that InceptionV3 has in reality significantly more parameters than ResNet18 or Densenet, our model choice for deployment is either of the latter as this would result in faster training and inference times.

In our results, we illustrated how this cascaded model can be used for solubility screening; however, different processes in material chemistry also rely on visual observation of solutions for multiphase materials systems.
Given the robustness and modularity of our method, we believe our model has huge potential to change visual assessment of samples in pharmaceutical and clean energy applications.



\section{Conclusion}
\label{sec:conclusion}
In the fields of materials discovery and pharmaceutical chemistry, the automatic screening of solutions is an important task and a staple procedure in various workflows.
We addressed this challenge of solubility screening in a real-world laboratory environment by introducing and evaluating SOLIS, a novel end-to-end cascaded model capable of determining whether a solute has dissolved in a particular solvent.
Specifically, we achieved this by combining image segmentation using a mask R-CNN for extracting the RoI with CNNs as image classifiers.
We demonstrated how our method possesses key advantages over existent ones, mainly in terms of not requiring human input and also by evaluating it on a dataset collected in an unmodified laboratory, without any inconvenient modifications to the laboratory.
Our results indicate that the proposed methodology can achieve excellent performance
on this challenging dataset, thus illustrating its potential to be used in autonomous laboratory experiments.

As the model has already been evaluated on images captured directly from the robotic manipulator, the evident next step is to deploy this model on the robot in a closed-loop material discovery workflow. 
The laboratory instruments and the robotic platform are already integrated in a robotic architecture for chemistry laboratories~\cite{Fakhruldeen2022} and hence this should be straightforward.
Other interesting research directions would be to look in more depth at using the system with organic solvents: this would require the workflow to be carried out in a fumehood. 
We believe our approach of training and evaluating the proposed model in real-world laboratory conditions will ease this.

In the future, additional refinements to the method could facilitate a more accurate estimation of the solubility, while still building on the same classifier outlined here; for example, in a titration like method involving several repeat measurements with the same solvent and solute combination. 
These repeats would employ smaller increments of solvent addition, longer equilibration times between liquid addition steps, and larger image buffers.
This may offer the potential for accurate measure of solubility limits, above and beyond the solubility thresholds that have been investigated here.


\bibliographystyle{ieeetr}
\bibliography{references}

\begin{thebibliography}{10}

\bibitem{Demir2019}
K.~A. Demir, G.~Döven, and B.~Sezen, ``Industry 5.0 and human-robot
  co-working,'' {\em Procedia Computer Science}, vol.~158, pp.~688--695, 2019.
\newblock 3rd World Conference on Technology, Innovation and Entrepreunership.

\bibitem{Bai2021}
J.~Bai, L.~Cao, S.~Mosbach, J.~Akroyd, A.~A. Lapkin, and M.~Kraft, ``From
  platform to knowledge graph: Evolution of laboratory automation,'' {\em JACS
  Au}, vol.~0, no.~0, p.~null, 0.

\bibitem{Shiri2021}
P.~Shiri, V.~Lai, T.~Zepel, D.~Griffin, J.~Reifman, S.~Clark, S.~Grunert, L.~P.
  Yunker, S.~Steiner, H.~Situ, F.~Yang, P.~L. Prieto, and J.~E. Hein,
  ``Automated solubility screening platform using computer vision,'' {\em
  iScience}, vol.~24, no.~3, p.~102176, 2021.

\bibitem{Inderwildi2020}
O.~Inderwildi, C.~Zhang, X.~Wang, and M.~Kraft, ``The impact of intelligent
  cyber-physical systems on the decarbonization of energy,'' {\em Energy
  Environ. Sci.}, vol.~13, pp.~744--771, 2020.

\bibitem{Petereit2011}
A.~Petereit and C.~Saal, ``What is the solubility of my compound? assessing
  solubility for pharmaceutical research and development compounds,'' {\em
  American Pharmaceutical Review}, vol.~14, pp.~68--73, 07 2011.

\bibitem{Eppel2020}
S.~Eppel, H.~Xu, M.~Bismuth, and A.~Aspuru-Guzik, ``Computer vision for
  recognition of materials and vessels in chemistry lab settings and the
  vector-labpics data set,'' {\em ACS Central Science}, vol.~6, no.~10,
  pp.~1743--1752, 2020.
\newblock PMID: 33145411.

\bibitem{Eppel2022}
S.~Eppel, H.~Xu, Y.~R. Wang, and A.~Aspuru-Guzik, ``Predicting 3d shapes{,}
  masks{,} and properties of materials inside transparent containers{,} using
  the transproteus cgi dataset,'' {\em Digital Discovery}, pp.~--, 2022.

\bibitem{Black2013}
S.~Black, L.~Dang, C.~Liu, and H.~Wei, ``On the measurement of solubility,''
  {\em Organic Process Research \& Development}, vol.~17, no.~3, pp.~486--492,
  2013.

\bibitem{daSilve2011}
A.~d. P.~M. da~Silva and J.~F. Cajaiba~da Silva, ``Determination of the adipic
  acid solubility curve in acetone by using atr-ftir and heat flow
  calorimetry,'' {\em Organic Process Research \& Development}, vol.~15, no.~4,
  pp.~893--897, 2011.

\bibitem{Burger2020}
B.~Burger, P.~M. Maffettone, V.~V. Gusev, C.~M. Aitchison, Y.~Bai, X.~Wang,
  X.~Li, B.~M. Alston, B.~Li, R.~Clowes, N.~Rankin, B.~Harris, R.~S. Sprick,
  and A.~I. Cooper, ``A mobile robotic chemist,'' {\em Nature}, vol.~583,
  pp.~237--241, Jul 2020.

\bibitem{He2017}
K.~He, G.~Gkioxari, P.~Doll{\'a}r, and R.~B. Girshick, ``Mask r-cnn,'' {\em
  2017 IEEE International Conference on Computer Vision (ICCV)},
  pp.~2980--2988, 2017.

\bibitem{He2016}
K.~He, X.~Zhang, S.~Ren, and J.~Sun, ``Deep residual learning for image
  recognition,'' {\em 2016 IEEE Conference on Computer Vision and Pattern
  Recognition (CVPR)}, pp.~770--778, 2016.

\bibitem{Simonyan15}
K.~Simonyan and A.~Zisserman, ``Very deep convolutional networks for
  large-scale image recognition,'' in {\em International Conference on Learning
  Representations}, 2015.

\bibitem{Szegedy2016}
C.~Szegedy, V.~Vanhoucke, S.~Ioffe, J.~Shlens, and Z.~Wojna, ``Rethinking the
  inception architecture for computer vision,'' in {\em 2016 IEEE Conference on
  Computer Vision and Pattern Recognition (CVPR)}, pp.~2818--2826, 2016.

\bibitem{Huang2017}
G.~Huang, Z.~Liu, L.~Van Der~Maaten, and K.~Q. Weinberger, ``Densely connected
  convolutional networks,'' in {\em 2017 IEEE Conference on Computer Vision and
  Pattern Recognition (CVPR)}, pp.~2261--2269, 2017.

\bibitem{Deng2009}
J.~Deng, W.~Dong, R.~Socher, L.-J. Li, K.~Li, and L.~Fei-Fei, ``Imagenet: A
  large-scale hierarchical image database,'' in {\em 2009 IEEE conference on
  computer vision and pattern recognition}, pp.~248--255, Ieee, 2009.

\bibitem{Fakhruldeen2022}
H.~Fakhruldeen, G.~Pizzuto, J.~Glawucki, and A.~I. Cooper, ``Archemist:
  Autonomous robotic chemistry system architecture,'' {\em IEEE International
  Conference on Robotics and Automation}, 2022.

\end{thebibliography}

\end{document}